\documentclass{article}
\usepackage[labelfont=it,textfont=it]{caption}
\usepackage[utf8]{inputenc}
\usepackage[T1]{fontenc}
\usepackage{multirow}
\usepackage{microtype}
\usepackage{graphicx}
\usepackage{float}
\usepackage{booktabs}
\usepackage{amsmath,amssymb,amsfonts}
\usepackage[letterpaper, margin=1.2in]{geometry}
\usepackage{times}
\usepackage{abstract}
\usepackage{titlesec}
\usepackage{hyperref}

\titlespacing{\section}{0pt}{10pt}{4pt}
\titlespacing{\subsection}{0pt}{8pt}{3pt}
\titlespacing{\subsubsection}{0pt}{6pt}{2pt}

\begin{document}
\thispagestyle{empty}
\hrule height 4pt
\vspace{6pt}
\begin{center}
    {\LARGE \bf Prediction Is Not Physics: Learning and Evaluating Conserved Quantities in Neural Simulators\par}
    
    \vspace{10pt}
    \hrule height 1pt
    \vspace{34pt}
    
    {\large \textbf{Andrew Bukowski}$^1$\textbf{, } \textbf{Aditya Kothari}$^1$\textbf{, } \textbf{Simba Shi}$^1$\textbf{, } \textbf{Ishir Rao}$^1$} \\
    \vspace{6pt}
    {\large $^1$Yale University} \\
\end{center}

\vspace{8pt}
\begin{center}
    {\large \bf Abstract}
\end{center}
\vspace{-8pt}

\begin{list}{}{
    \leftmargin=0.7in
    \rightmargin=0.7in
}
\item \small \noindent
A diffusion model trained on Hamiltonian trajectories can achieve rollout MSE near $10^{-3}$, but the standard deviation of its energy over time is measured between $7{,}500$ and $36{,}000$ times larger than the ground-truth energy standard deviation, indicating a failure to preserve conservation laws. This gap motivates our central question of whether neural networks can learn or select globally conserved quantities from physical trajectories. We investigate this across three Hamiltonian systems: projectile motion, pendulum, and spring-mass. We use a structured $T(v)+V(q)$ energy model, a black-box Conservation Discovery Network (CDN), a polynomial CDN, and a conditional diffusion baseline. The structured network reaches $R^2 \geq 0.9999$ against analytical energy on clean data, while the black-box CDN reaches $R^2 \geq 0.996$ when trained with temporal consistency plus a small alignment loss to analytical energy at $t{=}0$ ($\lambda_{\text{align}}{=}0.2$). With $\lambda_{\text{align}}{=}0$, CDN Pearson $R^2$ collapses on pendulum and spring-mass ($< 10^{-3}$), showing that temporal consistency alone is not enough to reliably identify the true energy. Under $1\%$ additive Gaussian noise, the CDN outperforms the structured model on the projectile and spring-mass systems, suggesting that the CDN may be more robust to noisy inputs in this setting. However, the polynomial CDN is sensitive to how it is trained: it achieves $R^2=0.78$ under a short training schedule on the pendulum system, but reaches $R^2=0.9998$ with more training time and data, regardless of whether noise is added.

\vspace{6pt}
\noindent
\textbf{Code:} \url{https://github.com/andrewbuko/452finalproject} \\
\textbf{Video:} \url{https://youtu.be/ypB4qmusx-I?si=VLAW5mOZk4Qg1lHg}

\end{list}
\vspace{5pt}

\section{Introduction}
\label{sec:intro}

Neural networks trained on data generated from physics simulations can make fairly accurate short-term predictions but, at the same time, may not obey fundamental physical laws. For example, they may allow objects to accelerate without a corresponding applied force, allow energy to drift over time, or fail to preserve conserved quantities because of the way the model is constructed.

Even when diffusion models are guided to some degree by learned conservation structure during sampling, they can still produce non-conservative rollouts. This indicates that low prediction error and soft invariant guidance are not enough to guarantee physically valid motion.

Instead of continuing to provide additional ``outside'' guidance to the generative model after it has been trained, we ask whether it is possible to achieve a different goal: to learn invariant properties directly from state observations of a physical system. If a model can identify an invariant from state trajectories, that invariant can be used to verify the model, guide new data generation, or help identify the physical laws that apply to the system.

To achieve this goal, we benchmark three different approaches to learning invariant values on three different Hamiltonian systems: projectile motion, the pendulum, and the spring-mass oscillator. Each system has a corresponding analytical invariant, which provides a way to assess the relative accuracy of the models according to conservation.

The three main questions we investigate are: Does accurate trajectory prediction imply conservation? How does model design affect the ability to recover conserved quantities? How can the training schedule influence the optimization landscape induced by conservation losses?

Our project makes four main contributions:

\begin{enumerate}
    \item We show that a neural network can approximate object motion while still violating conservation of energy. Diffusion-generated trajectories have within-trajectory energy standard deviation that is $7{,}500$--$36{,}000\times$ larger than the corresponding ground-truth trajectories.

    \item We evaluate conservation loss across three invariant-learning models: a black-box CDN, a polynomial CDN, and a structured energy network. This allows us to compare how different model designs affect the recovery of physically meaningful conserved quantities.

    \item We show that the polynomial CDN's pendulum $R^2$ increases from $0.78$ to $0.9998$ when trained with a longer schedule and more data, with or without input noise, suggesting that the model can remain in poor solutions under limited training budget.

    \item We show that the structured network's clean-data advantage narrows under $1\%$ state noise, with the black-box CDN outperforming the structured energy network on two of the three systems.
\end{enumerate}

\section{Related Work}
\label{sec:related}

Discovering physical laws from data has primarily focused on finding human-interpretable equations through techniques such as symbolic regression. Schmidt and Lipson~\cite{schmidt2009distilling} search directly for closed-form equations, while SINDy~\cite{brunton2016sindy} assumes the dynamics can be modeled using a limited number of terms from a known library. This work has been furthered by using new variables learned with an autoencoder~\cite{champion2019data} and by discovering mathematical equations through searches over expression trees, as in PySR~\cite{cranmer2023interpretable} and AI Feynman~\cite{udrescu2020ai}. These methods are most successful when the relevant variables are available and the search space is known to contain the anticipated functional form, but our setting asks whether a single conserved quantity can be learned directly from trajectories before trying to turn it into an explicit formula.

Building on Liu and Tegmark~\cite{liu2021machine}, we learn a conserved scalar quantity from observed trajectories, while treating closed-form recovery as a separate diagnostic rather than the main discovery task. Additionally, in this work we formulate a loss that encourages the learned quantity to remain constant along each trajectory, as well as a variance term that prevents the model from outputting the same value everywhere. When energy labels are available, we include a weak standardized alignment loss so that the learned value matches analytical energy up to scale and shift. We then determine whether approaches based on Conservation Discovery Networks can compare with models that have known physical structure, and whether learned invariant guidance is sufficient to make a diffusion model conserve energy while generating subsequent states.

Hamiltonian Neural Networks~\cite{greydanus2019hamiltonian} and Lagrangian Neural Networks~\cite{cranmer2020lagrangian} build principles based on energy or action into neural dynamics, often assuming that the energy of a dynamic system can be divided into kinetic energy $T(v)$ and potential energy $V(q)$. The structured energy network has a similar form, but serves as a baseline for testing rather than as the main object of study. While Neural ODEs~\cite{chen2018neural} provide a broad mathematical description of continuous-time systems, they do not impose conservation by default. In addition to the above works, diffusion and score-based models~\cite{ho2020denoising,song2020score} provide useful baselines for trajectory generation, and energy-guided sampling techniques~\cite{du2023reduce} show how learned constraints can be used during generation. Our results indicate that soft invariant guidance does not guarantee that generated trajectories conserve energy over time.
\section{Methods}
\label{sec:methods}

\subsection{Environments and Data Generation}

We evaluate three classical Hamiltonian systems covering both separable and nonlinear energies.

Projectile motion: state is $s=[x,y,v_x,v_y]$, conserved energy is
$E=\tfrac{1}{2}(v_x^2+v_y^2)+gy$, $g=9.81$. We generate projectile trajectories from the exact closed-form equations of motion, so the states are computed directly rather than approximated numerically.

Nonlinear pendulum: state is $s=[\theta,\omega]$, Hamiltonian is
$H=\tfrac{1}{2}\omega^2-g\cos\theta$ for unit mass and length. We sample initial conditions from $\theta_0\in[-\pi/2,\pi/2]$ and $\omega_0\in[-3,3]$, covering a range of libration energies, where the pendulum swings back and forth rather than rotating fully over the top. We simulate the pendulum using adaptive RK45, which automatically adjusts its step size to control numerical error, with relative and absolute error tolerances $\text{rtol}=10^{-12}$ and $\text{atol}=10^{-14}$.

Spring-mass system: state is $s=[x,v]$, energy is
$E=\tfrac{1}{2}kx^2+\tfrac{1}{2}v^2$ where $k=10$. Like the projectile motion, we use the exact closed-form solution, so the trajectory is computed directly from the equations of motion.

For each system, trajectories have length $T=200$ with physical timestep $\Delta t=0.005$ in the default pipeline. The default pipeline generates up to $10^6$ projectile trajectories and up to $5\times 10^5$ pendulum and spring-mass trajectories, using a 90/10 train-validation split with random seed 42. The CDN is trained on min-max-normalized states, the diffusion model is trained on standardized states, and the polynomial and structured energy models are trained on raw states. In noise experiments, additive Gaussian noise is applied before normalization and before the train-validation split. As a sanity check, we measure true energy drift along each ground-truth trajectory and find it is only $10^{-5}$ to $10^{-7}$. Therefore, larger conservation errors observed later can be attributed to the learned models rather than the simulated data.

\subsection{Conservation Loss}

All invariant-learning models are trained with an objective that has two goals: the learned quantity should stay nearly constant along each trajectory, but it should not collapse to the same value for every input. We write this loss as
\begin{align}
\mathcal{L} &= \underbrace{\frac{1}{B(T-1)}\sum_{i,t}\bigl(f(s_{i,t+1}) - f(s_{i,t})\bigr)^2}_{\text{temporal consistency}}
+ \lambda_{\text{var}} \cdot \underbrace{\max\!\bigl(0,\, \varepsilon - \operatorname{Var}_i[f(s_{i,0})]\bigr)}_{\text{variance hinge}} .
\label{eq:loss}
\end{align}
The first term encourages $f$ to have the same value at consecutive time steps in the same trajectory, pushing the model to learn something conserved over time. The second term prevents the trivial solution where $f$ outputs the same constant for every input by requiring $f(s_{i,0})$ to vary across trajectories.

For models trained with weak energy supervision, we also include a standardized alignment term at $t=0$:
\begin{equation}
\mathcal{L}_{\text{align}} = \frac{1}{B}\sum_i \left(\frac{f(s_{i,0})-\mu_f}{\sigma_f} - \frac{E(s_{i,0})-\mu_E}{\sigma_E}\right)^2 .
\label{eq:align}
\end{equation}
Here, $\mu_f,\sigma_f$ are the batch mean and standard deviation of the learned invariant values, while $\mu_E,\sigma_E$ are the batch mean and standard deviation of the analytical energy values. This term encourages the learned invariant to match the true energy up to scale and offset, helping select the physically meaningful conserved quantity. In our experiments, CDN-Conservation excludes this alignment term to test whether temporal consistency and non-collapse alone can learn a conserved scalar from trajectories, while CDN-Aligned, the polynomial CDN, and the structured energy network include it to help select the analytical Hamiltonian.

\subsection{Model Architectures}

\begin{figure}[H]
\centering
\includegraphics[width=0.85\linewidth]{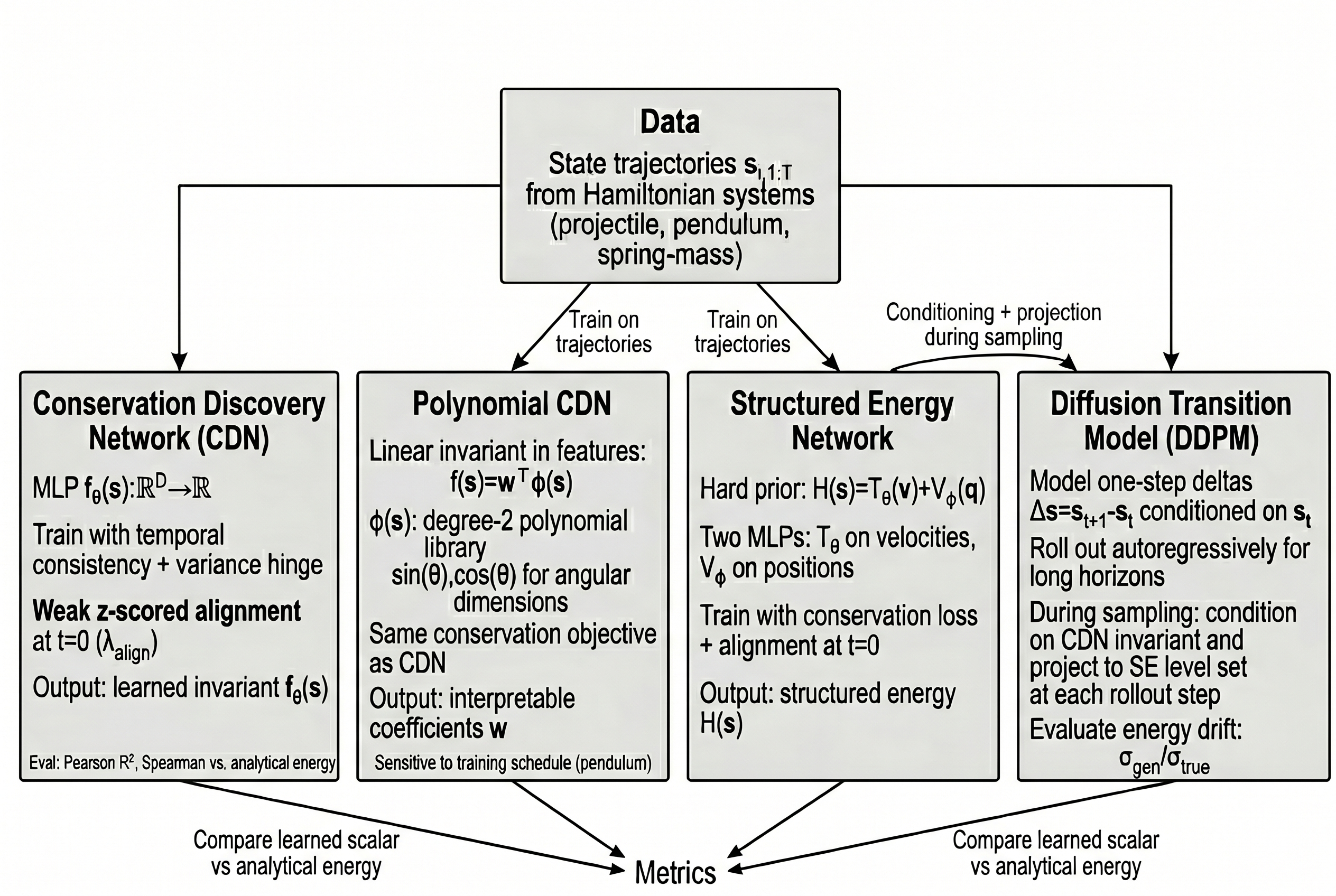}
\caption{Model schematic.}
\label{fig:schematic}
\end{figure}

The black-box CDN is an MLP with four hidden Linear--SiLU blocks of hidden dimension 256, mapping $s \in \mathbb{R}^D$ to a scalar invariant $f(s) \in \mathbb{R}$ with no imposed physical structure. It is trained on min-max-normalized states using the temporal consistency loss and variance-hinge regularizer in Equation~\ref{eq:loss}. We evaluate two variants: CDN-Conservation, which uses only the conservation objective, and CDN-Aligned, which additionally uses the standardized energy-alignment term in Equation~\ref{eq:align} with $\lambda_{\text{align}}=0.2$. CDN-Conservation tests whether an invariant can be learned from trajectories alone, while CDN-Aligned tests how much performance improves when the model is given weak analytical energy supervision.

The polynomial CDN is a linear invariant model
\[
f(s)=\mathbf{w}^{\top}\boldsymbol{\phi}(s),
\]
where $\boldsymbol{\phi}(s)$ expands the state into a library of simple candidate features. This library includes all degree-2 polynomial terms, along with trigonometric features for angular variables such as $\sin\theta$ and $\cos\theta$. The model then learns a weighted combination of these features as the conserved quantity. We train this model on raw, unnormalized states so that the learned weights can still be interpreted as physical coefficients. Training uses the same conservation and alignment losses, a 200-epoch linear warmup, CosineAnnealingWarmRestarts scheduling, $\lambda_{\text{var}}=1.0$, and $\lambda_{\text{align}}=0.2$.

The structured energy network decomposes the learned scalar as
\[
H(s)=T_\theta(v)+V_\phi(q),
\]
where $T_\theta$ and $V_\phi$ are separate two-hidden-layer MLPs with hidden dimension 128. This architecture builds in the usual mechanical energy structure: one network learns the velocity-dependent part, analogous to kinetic energy, and the other learns the position-dependent part, analogous to potential energy. In other words, the model is not allowed to learn an arbitrary function of the full state; it must represent the conserved quantity as a sum of a velocity-only term and a position-only term. The model is trained on raw states with the same conservation and alignment losses.

The diffusion transition model is a DDPM that learns to generate one-step changes in state,
\[
\Delta s=s_{t+1}-s_t,
\]
conditioned on the current state $s_t$. Rather than directly predicting the next state, the model learns a distribution over how the state changes from one time step to the next. The denoising network is a three-hidden-layer MLP with hidden dimension 256 and sinusoidal embeddings for the diffusion timestep. At inference, we generate full trajectories autoregressively: starting from an initial state, we sample a state change, update the state, and repeat this process forward in time.

During sampling, we guide the diffusion rollout using learned conservation structure. Specifically, the diffusion model is conditioned on the CDN invariant evaluated at the current state. After each sampled diffusion step, we apply, by default, one first-order projection step toward the structured energy network's initial energy level set:
\[
s \leftarrow s - \bigl(H_{\mathrm{SE}}(s)-H_0\bigr)
\frac{\nabla H_{\mathrm{SE}}(s)}{\|\nabla H_{\mathrm{SE}}(s)\|^2+\varepsilon},
\qquad
H_0=H_{\mathrm{SE}}(s_0),
\]
with $\varepsilon=10^{-8}$. This is a soft learned-energy correction rather than an exact analytical conservation constraint. This lets us test whether learned invariant guidance is sufficient to make diffusion-generated trajectories physically conservative.

\subsection{Symbolic Regression}

We apply STLSQ sparse regression over a polynomial-plus-trigonometric feature library, using the analytical energy as the regression target. In this setting, STLSQ tries to express the known energy function as a sparse weighted combination of candidate terms, such as polynomial terms and trigonometric terms. We discuss these results in Section~\ref{sec:results}.

This experiment confirms that our feature library is expressive enough to represent the true conserved quantities. However, it should not be interpreted as unsupervised discovery. The regression target is the known analytical energy, and the feature library already contains the correct terms needed to recover it. Thus, this experiment mainly checks that the symbolic regression pipeline can recover the right expression when given the right target and basis. We also implement PySR for tree-based symbolic search, but we do not include PySR results in the experiments below.

\section{Experiments and Results}
\label{sec:results}

We evaluate the models along four axes: alignment with analytical energy, sensitivity to training schedule, robustness to state noise, and the gap between rollout accuracy and global conservation. We compute correlations on trajectories not seen during training, and for the corresponding system, we compare the model's learned scalar invariant with the analytical Hamiltonian.

\subsection{Invariant Learning on Noise-Free Trajectories}

In Table~\ref{tab:main}, we record the $R^2$ and Spearman rank correlation between the main learned invariants and the noise-free trajectory analytical energy. $R^2$ is the square of Pearson correlation, invariant to affine transformations of learned outputs. This method is suitable given that all models are able to learn, without penalty, any affine reparameterization of the true energy.

\begin{table}[H]
\centering
\caption{Invariant quality on noise-free trajectories. Polynomial uses the short schedule; see Table~\ref{tab:schedule}. CDN-Conservation is discussed separately as the no-alignment ablation.}
\label{tab:main}
\small
\begin{tabular}{lcccccc}
\toprule
\multirow{2}{*}{\textbf{Model}} & \multicolumn{2}{c}{\textbf{Projectile}} & \multicolumn{2}{c}{\textbf{Pendulum}} & \multicolumn{2}{c}{\textbf{Spring-Mass}} \\
\cmidrule(lr){2-3}\cmidrule(lr){4-5}\cmidrule(lr){6-7}
 & $R^2$ & Sp. & $R^2$ & Sp. & $R^2$ & Sp. \\
\midrule
CDN-Aligned    & 0.997 & 0.998 & 0.997 & 0.9999 & 0.996 & 0.9997 \\
Polynomial     & 0.996 & 0.996 & 0.781 & 0.930  & 0.956 & 0.986  \\
SE Network     & \textbf{0.9999} & \textbf{0.9999} & \textbf{0.9999} & \textbf{0.9999} & \textbf{0.9999} & \textbf{1.0000} \\
\bottomrule
\end{tabular}
\end{table}

\begin{table}[H]
\centering
\caption{CDN-Conservation ablation without energy alignment. Pearson $R^2$ compares the learned scalar invariant to analytical energy.}
\label{tab:cdn_noalign}
\small
\begin{tabular}{lccc}
\toprule
\textbf{Model} & \textbf{Projectile} & \textbf{Pendulum} & \textbf{Spring-Mass} \\
\midrule
CDN-Conservation ($\lambda_{\text{align}}=0$) & 0.711 & $5.6\times10^{-4}$ & $2.0\times10^{-4}$ \\
\bottomrule
\end{tabular}
\end{table}

The structured energy network performs best across all systems, reaching $R^2 \geq 0.9999$. This is partly because the architectural prior matches the tested systems, since all three environments are exactly of the $T(v)+V(q)$ form. The black-box CDN-Aligned model has $R^2 \geq 0.996$ and Spearman $\geq 0.998$ on the three systems, showing that the conservation objective can recover a quantity representative of the true energy value, provided that it is paired with weak standardized energy alignment at $t{=}0$. The no-alignment ablation in Table~\ref{tab:cdn_noalign} shows that with $\lambda_{\text{align}}{=}0$, CDN Pearson $R^2$ falls to $5.6 \times 10^{-4}$ on pendulum, $2.0 \times 10^{-4}$ on spring-mass, and $0.711$ on projectile. This shows that the conservation objective alone is underdetermined and does not reliably select the analytical Hamiltonian.

\subsection{Polynomial-CDN: Schedule Sensitivity and the Conservation-Loss Landscape}
\label{sec:poly_brittle}

The Polynomial-CDN shows the largest schedule sensitivity among our models. On pendulum, it achieves $R^2 = 0.781$ on short schedule (256 epochs, 20K trajectories), and on long schedule it reaches $R^2 = 0.9998$.

The true pendulum Hamiltonian $\tfrac{1}{2}\omega^2 - 9.81\cos\theta$ lies in the polynomial-plus-trigonometric basis, and STLSQ regression over the same basis recovers exact coefficients given oracle energy targets (Table~\ref{tab:sindy}). The difficulty is one of optimization rather than expressivity.

In training, the mechanism is visible in the learned coefficients. In the short schedule, the polynomial converges to a solution dominated by the $\omega^2$ velocity term. This captures kinetic energy but misses the $\cos\theta$ potential term, so it acts as a partial proxy for the Hamiltonian rather than recovering the full conserved quantity. Near $\theta \approx 0$, the potential term varies approximately quadratically since $\cos\theta \approx 1 - \tfrac{1}{2}\theta^2$, which helps explain why an incomplete solution can still achieve moderate $R^2$ on much of the training distribution. With the longer schedule, the optimizer has enough epochs and data to move beyond this poor solution and recover the full $\cos\theta$ dependence. Most significantly, the final $R^2 = 0.9998$ is identical at $\sigma = 0$ and $\sigma = 0.01$ input noise, indicating that the extended training regime, rather than input perturbation, causes the improvement.

\begin{table}[H]
\centering
\caption{Schedule sensitivity of the polynomial-CDN on pendulum.}
\label{tab:schedule}
\small
\begin{tabular}{lccc}
\toprule
\textbf{Schedule} & \textbf{Epochs / Data} & $\boldsymbol{\sigma{=}0}$ & $\boldsymbol{\sigma{=}0.01}$ \\
\midrule
Short   & 256 ep / 20K trajs   & 0.781  & --     \\
Long    & 512 ep / 1M trajs    & 0.9998 & 0.9998 \\
\bottomrule
\end{tabular}
\end{table}

A related sensitivity affects the spring-mass system ($R^2$: $0.956 \to 1.000$ under the extended schedule), while the projectile is stable ($0.996$). Practitioners using conservation losses for symbolic discovery should consider that training-regime choices, including total epochs, learning-rate policy, and dataset size, can dominate model capacity as the binding constraint on recovery quality.

\subsection{Symbolic Regression via SINDy}

\begin{table}[H]
\centering
\caption{Equations recovered by STLSQ sparse regression.}
\label{tab:sindy}
\small
\begin{tabular}{lll}
\toprule
\textbf{System} & \textbf{Recovered} & \textbf{Ground Truth} \\
\midrule
Projectile  & $9.81\,y + 0.50\,v_x^2 + 0.50\,v_y^2$ & $gy + \tfrac{1}{2}v_x^2 + \tfrac{1}{2}v_y^2$ \\[3pt]
Pendulum    & $0.50\,\omega^2 - 9.81\cos\theta$      & $\tfrac{1}{2}\omega^2 - g\cos\theta$ \\[3pt]
Spring-Mass & $5.00\,x^2 + 0.50\,v^2$               & $\tfrac{1}{2}kx^2 + \tfrac{1}{2}v^2$ \\
\bottomrule
\end{tabular}
\end{table}

STLSQ on a polynomial library fit to analytical energy values recovers exact symbolic expressions across all three environments (Table~\ref{tab:sindy}). This should not be interpreted as unsupervised equation discovery. The regression uses the analytical energy as its target, and the basis is pre-populated with the correct monomial terms. This result is best interpreted as a pipeline and expressivity check rather than unsupervised discovery. Genuine symbolic discovery would require running STLSQ on the CDN's learned invariant values instead of the analytical energy, then checking whether the recovered expression matches the true conservation law, a direction we leave to future work.

\subsection{Diffusion vs.\ Conservation}

\begin{table}[H]
\centering
\caption{Diffusion transition model rollout accuracy vs.\ energy conservation.}
\label{tab:diffusion}
\small
\begin{tabular}{lcccc}
\toprule
\textbf{System} & \textbf{Rollout MSE} & $\boldsymbol{\sigma_{\text{true}}}$ & $\boldsymbol{\sigma_{\text{gen}}}$ & \textbf{Ratio} \\
\midrule
Projectile  & $1.8 \times 10^{-3}$ & $1.5 \times 10^{-5}$ & $1.1 \times 10^{-1}$ & $\mathbf{7{,}500\times}$ \\
Pendulum    & $3.4 \times 10^{-4}$ & $6.3 \times 10^{-7}$ & $2.3 \times 10^{-2}$ & $\mathbf{36{,}000\times}$ \\
Spring-Mass & $6.4 \times 10^{-3}$ & $1.1 \times 10^{-6}$ & $2.5 \times 10^{-2}$ & $\mathbf{23{,}000\times}$ \\
\bottomrule
\end{tabular}
\end{table}

Table~\ref{tab:diffusion} presents the central empirical result. The diffusion model achieves rollout MSE as low as $3.4 \times 10^{-4}$ on the pendulum, yet the within-trajectory energy standard deviation of its generated rollouts is $36{,}000\times$ larger than in ground-truth trajectories. The ratio is similarly extreme on the other two systems, reaching $7{,}500\times$ on projectile and $23{,}000\times$ on spring-mass. This result holds despite the diffusion model being conditioned on the CDN invariant and softly guided toward the structured energy network's learned energy level set at each sampling step. Because the guidance is a soft gradient correction toward the learned energy level set rather than an exact analytical constraint, cumulative autoregressive error can still produce substantial analytical-energy drift. Even a model trained to minimize denoising loss and softly guided by learned invariant structure can achieve low rollout error while still failing to enforce global conservation. This directly implies that trajectory-level accuracy metrics such as rollout MSE are insufficient benchmarks for physical simulation models, and that conservation-based metrics must be evaluated as a separate axis of quality.

\subsection{Noise Robustness}

\begin{table}[H]
\centering
\caption{$R^2$ under 1\% Gaussian state noise.}
\label{tab:noise}
\small
\begin{tabular}{lccc}
\toprule
\textbf{Model} & \textbf{Projectile} & \textbf{Pendulum} & \textbf{Spring-Mass} \\
\midrule
CDN-Aligned & \textbf{0.993} & 0.978 & \textbf{0.993} \\
SE Network  & 0.987          & \textbf{0.986} & 0.987 \\
\bottomrule
\end{tabular}
\end{table}

Under 1\% additive Gaussian state noise (Table~\ref{tab:noise}), the SE network's $R^2$ drops by approximately $0.013$ uniformly across all three systems. The CDN-Aligned model drops slightly on projectile ($0.997 \to 0.993$) and on pendulum ($0.996 \to 0.978$), while remaining essentially flat on spring-mass ($0.996 \to 0.993$). The net effect is that on projectile and spring-mass the CDN-Aligned model now exceeds the SE network, reversing the clean-data ordering, while on the pendulum the SE network retains a small advantage. The architectural prior that confers near-perfect performance on noise-free data appears to be more brittle under perturbation. Forcing the $T(v)+V(q)$ decomposition on noisy states requires the sub-networks to fit noise-corrupted position and velocity signals separately, whereas the unconstrained CDN can learn any robust function of the full state. This is a preliminary observation from a single seed per condition and warrants systematic study with multiple noise levels and seeds.

\section{Discussion}

\label{sec:discussion}

Structural inductive bias can be beneficial when correct; however, for hard-coded decompositions of $T(v)+V(q)$, the function class used in all three systems is constrained and produces nearly identical $R^2$ values. This is partially by definition of the three systems we analyzed. The CDN-Aligned model produced $R^2 \geq 0.996$ without structural constraints but with weak standardized energy alignment, making it more flexible architecturally than the structured energy network. As a learning target, conservation with weak standardized energy alignment at $t=0$ can produce a strongly monotonic quantity relative to the corresponding true energy without directly committing to the particular physical form of that energy. Outside of the $T(v)+V(q)$ function class, such as velocity-dependent potentials, dissipative mechanics, or Lorentz-force mechanics, the benefit of CDN flexibility and the limits of SE network priors would be substantial.

Learning physically consistent trajectories can also be achieved through Hamiltonian Neural Networks that encourage conservation by computing the dynamics from a learned scalar $H$ through Hamilton's equations, or through symplectic integrators, such as Störmer-Verlet-type algorithms, that preserve Hamiltonian structure through their numerical integration update rules rather than through the learning objective. Although both techniques provide stronger conservation structure than the CDN method, both require the system to either know or assume a predefined form for the dynamics, while the CDN method is able to operate on observed trajectories without requiring the user to know the physics of the system other than the assumption that some conserved quantity exists.

The polynomial-CDN curve fit yielded some non-intuitive insights into the conservation objective. While the loss function presented in Equation~\ref{eq:loss} is relatively simple to understand, the landscape generated by the loss is much more complicated. After a short period of training, the polynomial solution converges to the kinetic-energy term ($\omega^2$), but not the potential-energy term ($\cos\theta$) represented in the Hamiltonian. Thus, the overall quantity learned is, at best, only a partial Hamiltonian. The variance hinge compounds this difficulty because it creates a flat-gradient region once $\operatorname{Var}_i[f(s_{i,0})]$ is greater than $\varepsilon$, removing signal when the consistency loss would otherwise continue refining the solution. The observations from the longer schedule, with more epochs and training data, show that the optimizer can move beyond this poor partial solution and recover the complete $\cos\theta$ dependence. This reflects that the training method was the operative variable rather than input perturbation. The fact that $R^2=0.9998$ for both $\sigma=0$ and $\sigma=0.01$ further validates this result.

The surprising exact coefficient recovery with STLSQ needs to be accurately represented. This method uses oracle targets and its basis contains the answer. For genuine symbolic discovery, STLSQ must be executed on the invariant values learned by the CDN, then compared to the actual conservation law to verify whether the recovered expression corresponds to the true law. Completing the CDN-SINDy pipeline is the next logical step.

The $\lambda_{\text{align}}=0$ collapse indicates that numerous fixed points exist for the conserved quantity, but only one corresponds to the physical Hamiltonian. The alignment term is used to select this solution, meaning that the CDN can learn temporally conserved scalar quantities, but requires weak supervision to determine which conserved scalar corresponds to analytical energy.

\section{Conclusion}

\label{sec:conclusion}

We show that trajectory-level prediction accuracy and physical conservation are distinct axes of model quality, and we provide baseline performance measures for several classes of models across three Hamiltonian systems.

The diffusion transition model produces low rollout mean squared error (MSE), but has an energy standard-deviation ratio of $7{,}500$ to $36{,}000$ times greater than ground truth, even when conditioned on a learned invariant and softly guided toward a learned energy level set. This demonstrates that local denoising objectives and soft conservation guidance are insufficient for correctly simulating physical systems.

The structured energy network, using the $T(v)+V(q)$ approach, yields near-perfect agreement with analytical energy when the prior used for training the model matches the system class; however, when the model is tested under $1\%$ state noise, this measured alignment advantage decreases.

The polynomial CDN's short training schedule produces an $R^2$ of $0.78$, while the long training schedule produces an $R^2$ of $0.9998$. This suggests that the conservation objective can lead to poor partial solutions under limited training budget, and that sufficient epochs and data are important for reliable recovery.

All experiments were done with a single seed per condition at the trajectory level. While many qualitative results appear similar across trajectory systems, repeating the experiment with multiple seeds and a systematic set of added state noise conditions is required to validate those results.

There are many areas to explore in the future through continued study and experimentation. A useful way to visualize the amount of drift in both the ground-truth and generated trajectories over time would be through direct time-series plots of the learned energy.

Lastly, a natural next step is stronger energy-guided DDIM sampling, where the learned invariant directly shapes each denoising step rather than only providing the current conditioning and soft post-step correction used here. This is motivated by the large difference between the energy standard deviations of the generated trajectory data and the ground-truth trajectory data in Table~\ref{tab:diffusion}.

\section{Acknowledgments}

\label{sec:acknowledgments}

This project was done as part of the \textit{Deep Learning Theory and Applications} (CPSC 4520) course at Yale University, taught by Professor Smita Krishnaswamy. We thank her for her valuable feedback on this work.

\end{document}